\documentclass[final,3p,times,twocolumn]{elsarticle}
\usepackage{lineno,hyperref}
\usepackage{multicol}
\usepackage{multirow}
\usepackage{amsmath}
\usepackage{wrapfig}
\usepackage{threeparttable}
\journal{Signal Processing: Image Communication}%

\begin{document}

\begin{frontmatter}



\title{Grafted Network for Person Re-Identification}

\author[mymainaddress]{Jiabao Wang\corref{correspondingauthor}}
\ead{jiabao\_1108@163.com}
\cortext[correspondingauthor]{Corresponding author}

\author[mymainaddress]{Yang Li}
\ead{solarleeon@outlook.com}

\author[mymainaddress]{Shanshan Jiao}
\ead{597241031@qq.com}

\author[mymainaddress]{Zhuang Miao}
\ead{emiao\_beyond@163.com}

\author[mymainaddress]{Rui Zhang}
\ead{3959966@qq.com}

\address[mymainaddress]{College of Command and Control Engineering, Army Engineering University of PLA, China}

\begin{abstract}
Convolutional neural networks have shown outstanding effectiveness in person re-identification (re-ID). However, the models always have large number of parameters and much computation for mobile application. In order to relieve this problem, we propose a novel grafted network (GraftedNet), which is designed by grafting a high-accuracy rootstock and a light-weighted scion. The rootstock is based on the former parts of ResNet-50 to provide a strong baseline, while the scion is a new designed module, composed of the latter parts of SqueezeNet, to compress the parameters. To extract more discriminative feature representation, a joint multi-level and part-based feature is proposed. In addition, to train GraftedNet efficiently, we propose an accompanying learning method, by adding an accompanying branch to train the model in training and removing it in testing for saving parameters and computation. On three public person re-ID benchmarks (Market1501, DukeMTMC-reID and CUHK03), the effectiveness of GraftedNet are evaluated and its components are analyzed. Experimental results show that the proposed GraftedNet achieves 93.02\%, 85.3\% and 76.2\% in Rank-1 and 81.6\%, 74.7\% and 71.6\% in mAP, with only 4.6M parameters.

\end{abstract}


\end{frontmatter}

%\linenumbers

\section{Introduction}
\label{intro}
In person re-identification (re-ID), convolutional neural network (CNN) is an effective method to extract features for person representation. Unfortunately, the current high-accuracy network, such as ResNet~\cite{DBLP:conf/cvpr/HeZRS16}, always has too many parameters for storage and is too time-consuming for computation. On the contrary, the light-weighted network, such as SqueezeNet~\cite{DBLP:journals/corr/IandolaMAHDK16}, usually has low accuracy even if it has fewer parameters and efficient calculation. The existing methods always develop new networks by designing light-weighted modules or high-accuracy modules. Different from these methods, in this paper we try to explore a new light-weighted and high-accuracy network by grafting two existing networks.

In botanical research area, grafting is a widely used effective method to cultivate new varieties. It combines two different plants, attaching one plant branch to another plant¡¯s stem, so that they can heal together and form an independent new individual. The grafted stem, called rootstock, becomes the root part of the plant after grafting, while the grafted branch, called scion, becomes the upper part of the plant after grafting. Grafting can maintain the excellent characteristics of different plants, enhance the resistance and adaptability. Inspired by the above idea, we try to develop a new grafted network (GraftedNet) by grafting a high-accuracy network and a light-weighted network.

For grafting, rootstock and scion are the two basic components. Rootstock provides water and inorganic nutrients to the scion. Therefore, the quality of rootstock plays an important role in the survival rate of grafting, the growth and development of grafted plants, as well as resistance and adaptability. In addition, an affinity between scion and rootstock is also the main factor affecting the survival of grafting, that is, the ability of scion and rootstock to combine with each other in terms of internal structure, physiology and heredity. As a result, we need to find a strong network as a rootstock to provide high-accuracy, and find a light-weighted network as a scion to provide few parameters. Furthermore, we also need to ensure the affinity between rootstock and scion.

For person re-ID, ResNet-50 is the widely used network and achieves high-accuracy performance, so it is the first choice for rootstock. However, if we directly use the whole ResNet-50 as rootstock, GraftedNet can't have few parameters and efficient computation. So we remove the latter parts of ResNet-50 and keep the former parts of ResNet-50 as a rootstock. After rootstock is selected, another important task is to find scion. The simplest method is to find it in the existing light-weighted networks, such as SqueezeNet~\cite{DBLP:journals/corr/IandolaMAHDK16}, MobileNet~\cite{DBLP:journals/corr/HowardZCKWWAA17}, ShuffleNet~\cite{DBLP:conf/cvpr/ZhangZLS18}. However, we can't directly graft the modules of the light-weighted networks with the rootstock, due to their different structures. As a result, we make some modification based on the latter parts of the existing light-weighted networks. Based on above modification, our GraftedNet can be constructed. To extract more discriminative feature, a joint multi-level and part-based feature is proposed. However, there is another important question: how to ensure that the new grafted network has high accuracy? According to our experiments, the grafted network can't achieve the high-accuracy if we train it with the simplest fine-tune strategy. To solve this problem, we propose a new accompanying learning method for training the grafted network. The final performance of our proposed grafted network achieves 93.02\% in Rank-1 and 81.58\% in mAP on Market1501 dataset, with only 4.6M parameters.

The contributions of this work are presented as follows:
\begin{itemize}
  \item A novel light-weighted and high-accuracy grafted network (GraftedNet) is proposed. To the best of our knowledge, GraftedNet is the first method which can achieves better performance than the original ResNet-50, with only 4.6M parameters.
  \item A joint multi-level and part-based feature is proposed for extracting more discriminative feature representation.
  \item An accompanying learning method is proposed for training GraftedNet. An accompanying branch is added to GraftedNet for supervised learning and can be removed in testing for saving parameters and computation.
  \item Experiments are conducted on the public Market1501, DukeMTMC-reID and CUHK03 datasets. The effectiveness of GraftedNet has been evaluated and its components are compared and analyzed.
\end{itemize}

\section{Related Work}
\subsection{Light-weighted Networks}
In the past few years, deep convolutional neural networks, such as ResNet~\cite{DBLP:conf/cvpr/HeZRS16} and VGGNet~\cite{DBLP:journals/corr/SimonyanZ14a}, have achieved remarkable results in various tasks in the field of computer vision. However, these networks are not suitable for deployment on mobile platforms and edge devices because of the cost of computation and storage. Therefore, researchers proposed several small light-weighted models, such as SqueezeNet~\cite{DBLP:journals/corr/IandolaMAHDK16}, MobileNet~\cite{DBLP:journals/corr/HowardZCKWWAA17}, ShuffleNet~\cite{DBLP:conf/cvpr/ZhangZLS18}.

SqueezeNet~\cite{DBLP:journals/corr/IandolaMAHDK16} is a compression model consists of a series of the special designed fire modules. The model just has 1.2M parameters for classifying ImageNet 1000 classes with 57.5 \% top-1 accuracy, which just equals that of AlexNet~\cite{DBLP:conf/nips/KrizhevskySH12}. The main reason is that the original version has no skip-connection which is proved very effective in ResNet~\cite{DBLP:conf/cvpr/HeZRS16}.

MobileNet~\cite{DBLP:journals/corr/HowardZCKWWAA17} is designed by dividing the standard convolution into depth-wise convolution and point-wise convolution, both of which can compress the model by several times. The model has 4.2M parameters for classifying ImageNet 1000 classes with 70.6 \% top-1 accuracy. The second version (MobileNet V2~\cite{DBLP:conf/cvpr/SandlerHZZC18}) uses inverted residual structure and linear bottlenecks to promote the performance, achieving 71.7\% top-1 accuracy with 3.4M parameters.

ShuffleNet~\cite{DBLP:conf/cvpr/ZhangZLS18} uses point-wise group convolution and channel shuffle to reduce computation cost and promote accuracy. Channel shuffle operation is used to help the information flowing across feature channels separated by group convolution. It achieves 65.9\% top-1 accuracy with 1.8M parameters. ShuffleNet V2~\cite{DBLP:conf/eccv/MaZZS18} provides a more effective network architecture, achieving 69.4\% top-1 accuracy with the same number of parameters.

Although these light-weighted networks have fewer parameters and efficient computation, there is a big gap in the mAP and Rank-1, comparing with high-accuracy networks, such as ResNet-50~\cite{DBLP:conf/cvpr/HeZRS16}.

\subsection{Distilling Learning}
Hinton et al.~\cite{DBLP:journals/corr/HintonVD15} proposes the distilling method for model compression by transferring knowledge from an ensemble or from a large regularized model into a smaller, distilled model. It firstly trains a big network in training and produces a small network in deployment. And the smaller network meets the requirements of low storage and high efficiency. Partially inspired by the idea, we propose a different architecture. Distillation use two independent networks, a large model and a small, distilled model, while our GraftedNet has a parameter-shared rootstock and two independent branches, one of which is an accompanying branch. Distillation transfers the knowledge by giving the predicted soft targets to distilled model, while our GraftedNet has the same truth label for both branches.

Model distillation can be treated as an effective technology for transferring knowledge from teacher model to student model. However, the teacher-student network~\cite{DBLP:journals/corr/abs-1902-00643} is widely used in semi-supervised learning. It aims at learning with limited labeled data and abundant unlabeled data. The teacher generates the targets for training the student. For GraftedNet, the accompanying branch can also be treated as a teacher, and the student is taught by updating the rootstock. The accompanying branch and the scion share the same parameters of the rootstock, which is different from the original distillation models.

Different from teacher-student network, mutual learning network~\cite{DBLP:conf/cvpr/ZhangXHL18} has two student sub-networks, rather than one-way students from teachers. Both student sub-networks are not pre-trained and can learn from each other at the same time, so as to solve the target task. In this work, the accompanying branch and the scion can be treated as an experienced senior and a freshman respectively. They are trained at the same time for the same classification task.

\section{GraftedNet}
\begin{figure}[!t]
\centering
\includegraphics[width=3.0in]{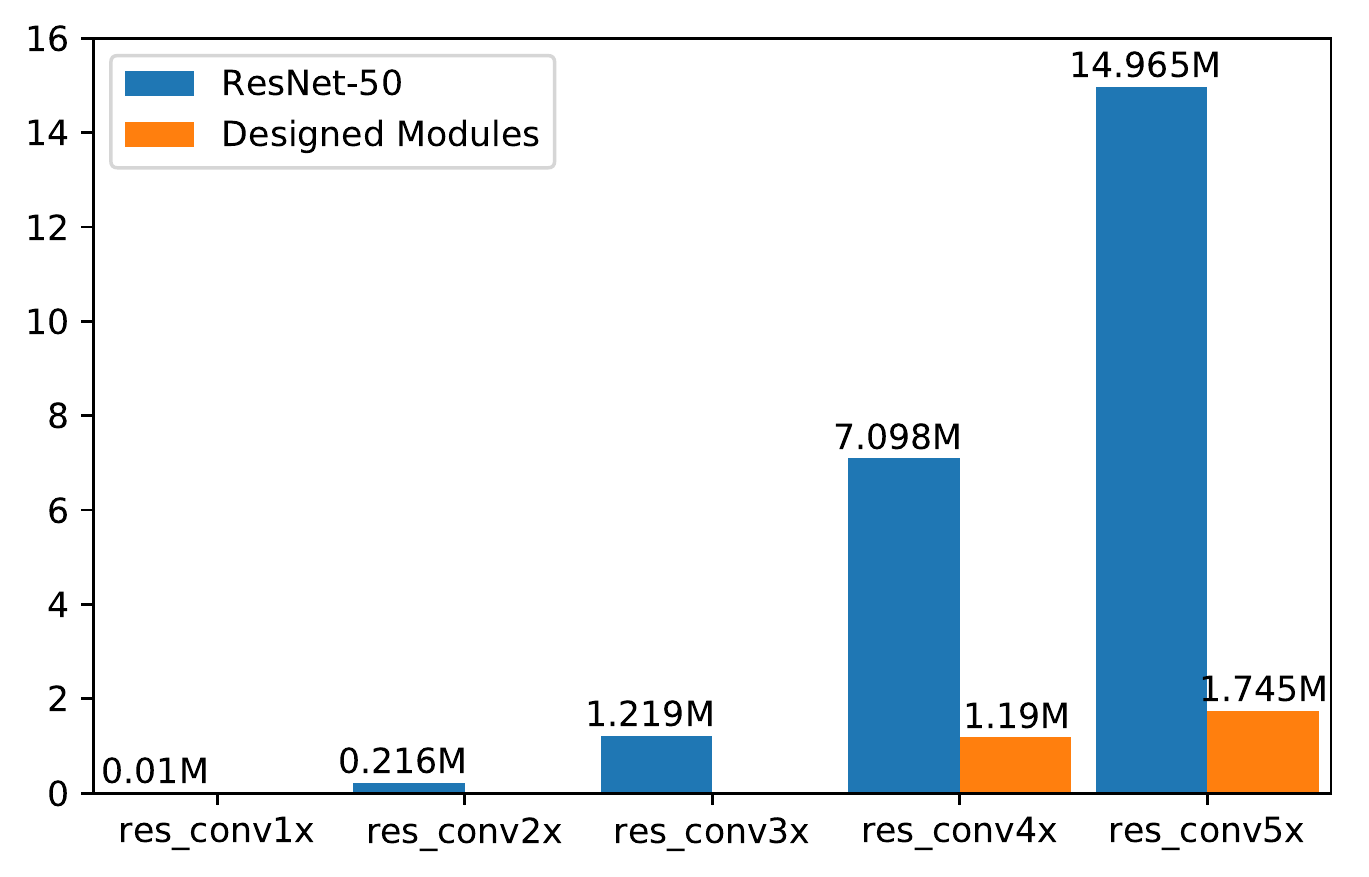}
\caption{The number of parameters of ResNet-50 and designed modules. The blue bar represents the number of parameters in each stage, while the brown bar represents the number of parameters of the new designed modules for scion.}
\label{fig1_ParamBar}
\end{figure}

As one of the best CNN method in person re-ID, MGN~\cite{DBLP:conf/mm/WangYCLZ18} has achieved 95.6\% in Rank-1 and 86.9\% in Mean Average Precision (mAP) on Market1501 dataset. It is designed based on ResNet-50 and boosts the performance by three same-structural and parameter-independent branches, each of which has about 22M parameters and there are nearly 69M parameters in total. So it is inappropriate for mobile application. To further analyze the number of parameters in ResNet-50, we divide the original ResNet-50 into five stages, noted as res\_conv1x, res\_conv2x, res\_conv3x, res\_conv4x and res\_conv5x, according to the feature map size. In each stage, there are several residual blocks, which can be indexed in a ``stage(number)+block(alphabet)'' manner, e.g. res\_conv5a for block 1 in stage 5. The number of parameters in each stage is shown in Figure \ref{fig1_ParamBar}, where blue bar represents the number of parameters in each stage.

From Figure \ref{fig1_ParamBar}, we can find that there are 22.063M parameters in res\_conv4x and res\_conv5x stages. For MGN, its three branches have the same structure of res\_conv4x and res\_conv5x, which hold most of the parameters of ResNet-50. That¡¯s why MGN has too many parameters. According to above analysis, most of the parameters in ResNet-50 come from res\_conv4x and res\_conv5x, while the number of parameters in first three stages is only 1.445M. At this point, we should remove res\_conv4x and res\_conv5x for saving parameters. So the rootstock is composed of the first three stages of ResNet-50. According to our experiments, we also find the first three stages have much better performance than the first two stages, with tolerable number of increased parameters. So the rootstock of grafted network is the first three stages of ResNet-50.

After rootstock is decided, scion can be selected from the existing light-weighted networks, such as SqueezeNet~\cite{DBLP:journals/corr/IandolaMAHDK16}, MobileNet~\cite{DBLP:journals/corr/HowardZCKWWAA17}, ShuffleNet~\cite{DBLP:conf/cvpr/ZhangZLS18}. To graft rootstock and scion, we need to do some modifications to affine both of them. For simplicity, we directly modify the structure of the latter stages of the existing light-weighted networks. The details of the scion are presented in section \ref{scion}. Through grafting technique, the model size can be compressed by grafting the former stages of ResNet-50 and the latter stages of a lighted-weighted network.

\subsection{Architecture}
\begin{figure*}[!t]
\centering
\includegraphics[width=5.5in]{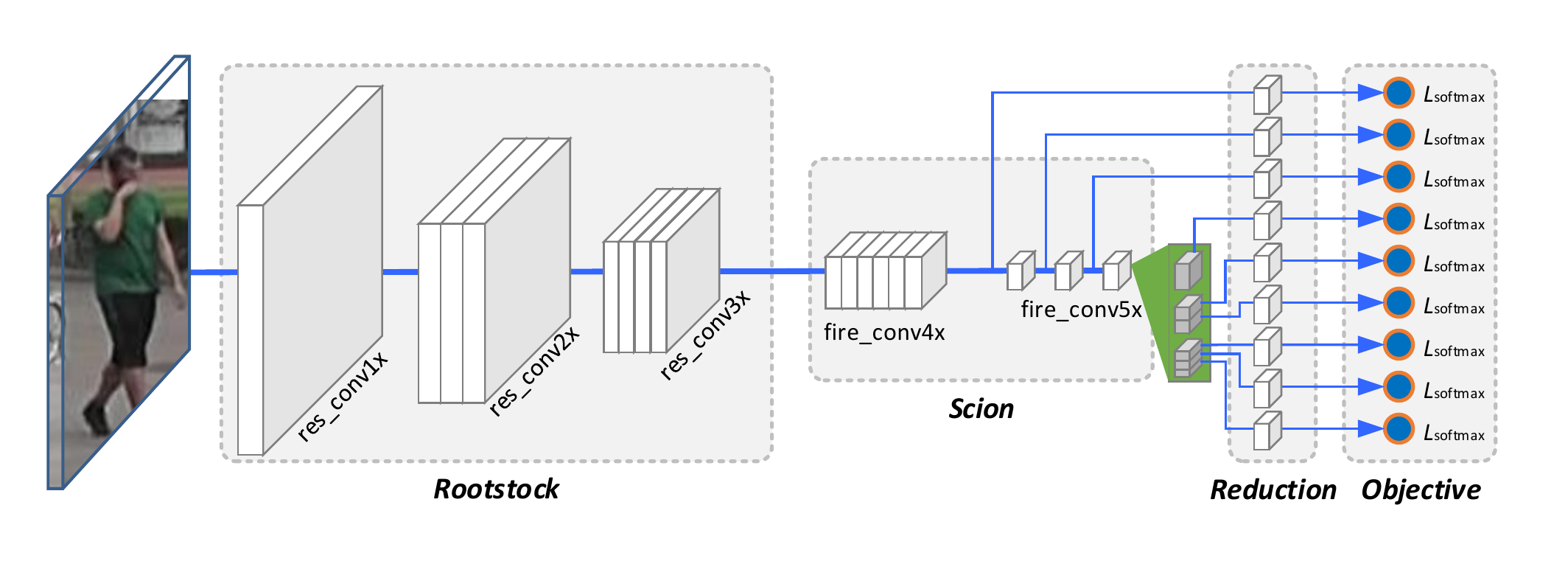}
\caption{The architecture of our proposed GraftedNet. \emph{Rootstock} is consist of the first three stages of ResNet-50, while \emph{Scion} is composed of new designed modules based on the fire block, which is consists of squeeze and expansion operations. \emph{Reduction} is used for extracting joint multi-level and part-based feature with low dimension. \emph{Objective} is used for supervised learning.}
\label{fig2_Architecture}
\end{figure*}

Figure \ref{fig2_Architecture} shows the architecture of our proposed GraftedNet, which is consists of four parts, \emph{Rootstock}, \emph{Scion}, \emph{Reduction} and \emph{Objective}.

\emph{Rootstock}: The rootstock of GraftedNet consists of res\_conv1x, res\_conv2x and res\_conv3x of ResNet-50. It inputs a person image and outputs a set of feature maps, which provides the low-level information to \emph{Scion}. It has low computational cost, with is only 1.445M parameters. Especially, it provides very effective feature maps when the parameter is initialized by ImageNet pre-trained model.

\emph{Scion}: It composed of two new designed modules, fire\_conv4x and fire\_conv5x. The two modules are designed based on the fire block, which consists of the squeeze and expansion operations. The details of the two modules are presented in the section \ref{scion}.

\emph{Reduction}: The output of fire\_conv5x (fire\_conv4x) is 768-channels (512-channels) feature map. After global average pooling (GAP) operation, the reserved feature is a 768-dims (512-dims) vector. To represent a person effectively, we decrease the feature to a low-dimensional space. \emph{Reduction} is composed of a 1$\times$1 group convolution, followed by a batch normalization and a leaky ReLU with the negative slope of 0.1. It reduces feature from 768-dims(512-dims) to 256-dims, which is only the 1/3 (1/2) of the original dimension. Furthermore, group convolution is used to decrease the parameters of \emph{Reduction}. When we set the number of group as 8 in group convolution, the parameters can be reduced by 1.491M with a degradation of mAP less than one percent. The details can be found in section \ref{group}.

\emph{Objective}: The objective is set at multiple high-level layers of GraftedNet. It comes from the outputs of fire\_conv4f, fire\_conv5a, fire\_conv5b, fire\_conv5c. Specially, inspired by MGN~\cite{DBLP:conf/mm/WangYCLZ18}, we divide the feature maps of fire\_conv5c into different parts vertically. The division has one part, two parts and three parts. For each feature produced by \emph{Reduction}, we use a 1$\times$1 convolution to transform the 256-dims feature into the number of person identities. The 9 softmax log-loss objectives are jointly used for classification.

\subsection{The Designed Scion}
\label{scion}
\begin{figure}[!t]
\centering
\includegraphics[width=3.0in]{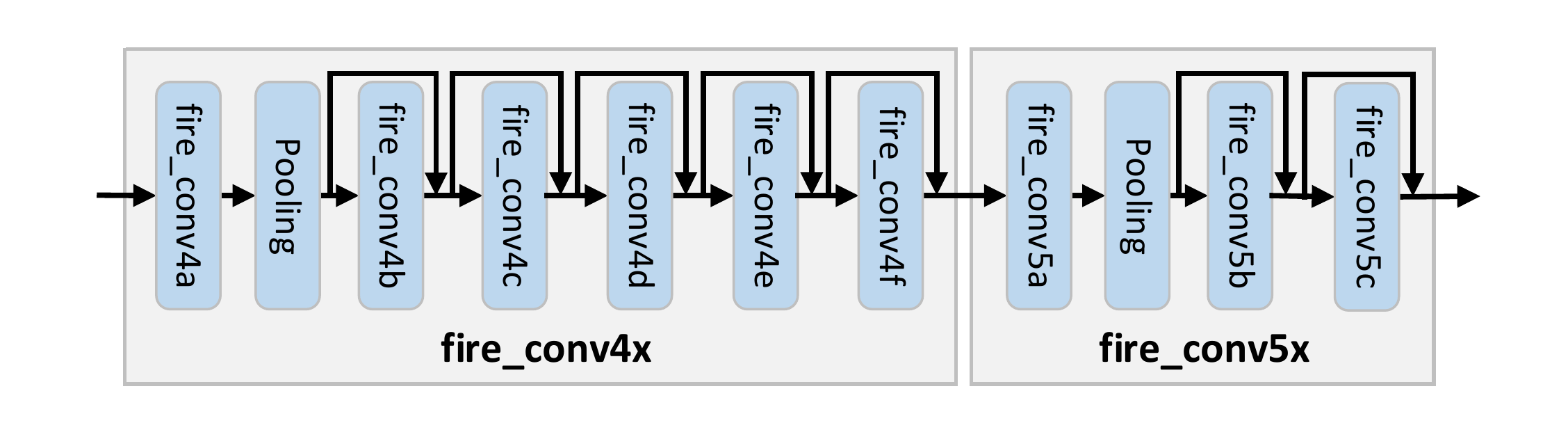}
\caption{The architecture of the designed scion.}
\label{fig3_Scion}
\end{figure}

To better fit the rootstock, we design the scion according to the structure of stage 4 and stage 5 of ResNet-50. The detail of the structure is shown in Figure \ref{fig3_Scion}.

Note that we add the skip-connection to all layers which have the same output size with the former layer. This can promote the performance greatly. To connect with the rootstock, the channels of the fire\_conv4a is 512, while the planes of squeeze operation is 64 for compress parameters, and the planes of expand operation 1$\times$1 and 3$\times$3 is 256 for the same channels. In fire\_conv5x, the channels is 768, and the planes of squeeze operation is 128, and the planes of expand operation 1$\times$1 and 3$\times$3 is 384 for the same channels. After fire\_conv4a and fire\_conv5a, we add a max-pooling operation with the kernel of size 3 and stride 2 to reduce the size of the feature maps.

\subsection{Joint Multi-level and Part-based Feature}
According to the part-based models, such as PCB-RPP~\cite{DBLP:conf/eccv/SunZYTW18}, the features extracted from local area of a person image has a strong ability to boost the accuracy for person re-ID. So we introduce the part-based features, but with the multi-scale description of the output feature map of res\_conv5c. The multi-scale idea is very similar to MGN~\cite{DBLP:conf/mm/WangYCLZ18}, but just for the same feature map, not for different feature maps of three branches. In addition, inspired by GoogleNet~\cite{DBLP:conf/cvpr/SzegedyLJSRAEVR15}, we set multiple objectives at multiple high-level layers. It is different from the existing methods, which always extract features from the output of res\_conv5c in ResNet-50. Except res\_conv5c, we extract the features from the layers of fire\_conv4f, fire\_conv5a, fire\_conv5b in GraftedNet. As a result, we get 9 features in total. For training and testing, we adopt \emph{Reduction} to extract the low-dimensional features from the high-dimensional feature maps. The dimension of the output of fire\_conv4f is $H\times W \times 1024$, where $H$ and $W$ refers the height and width of the tensor of the feature maps. The dimension of the output of fire\_conv5a, fire\_conv5b and fire\_conv5c is $H\times W \times2048$.

In training, the processing of \emph{Reduction} is first to obtain a $1\times 1\times 1024(2048)$  feature vector by the global average pooling operation. Then, we use the linear transformation to compress the feature form a high dimension to a low dimension, which is a relative same value. Finally, we use a softmax log-loss to supervise the training of the GraftedNet.

If we have a batch of person images$\{{{I}_{i}}\}_{i=1}^{B}$, the corresponding feature ${{\mathbf{f}}^{k}}({{I}_{i}})$ can be obtained from the $k$-th \emph{Reduction}. The softmax log-loss is then computed from feature ${{\mathbf{f}}^{k}}({{I}_{i}})$ and its truth label ${{y}_{i}}$. Each objective corresponds to one loss, which has the form of:
\begin{equation}\label{eq1}
\begin{aligned}
L_{softmax}^{k}=-\sum\limits_{i=1}^{B}{\log \frac{\exp ({{(\mathbf{W}_{{{y}_{i}}}^{k})}^{T}}{{\mathbf{f}}^{k}}({{I}_{i}})+b_{{{y}_{i}}}^{k})}{\sum\nolimits_{j=1}^{C}{\exp ({{(\mathbf{W}_{j}^{k})}^{T}}{{\mathbf{f}}^{k}}({{I}_{i}})+{{b}_{j}}^{k})}}}
\end{aligned}
\end{equation}
where $B$ is the mini-batch size, $C$ is the number of classes, and $\mathbf{W}_{j}^{k}$ and $b_{j}^{k}$ are the parameters of the $k$-th objective to learn.

The classification joint objective is the summation of all 9 objectives,
\begin{equation}\label{eq2}
\begin{aligned}
{{L}_{joint}}=\sum\limits_{k=1}^{9}{L_{softmax}^{k}}
\end{aligned}
\end{equation}
In testing, given a query or gallery person image ${{I}_{t}}$, its representation can be obtained by concatenating the 9 features ${{\mathbf{f}}^{k}}({{I}_{t}}),k=1,2,...,9$
\begin{equation}\label{eq3}
\begin{aligned}
\mathbf{f}({{I}_{t}})=[{{\mathbf{f}}^{1}}({{I}_{t}}),{{\mathbf{f}}^{2}}({{I}_{t}}),...,{{\mathbf{f}}^{9}}({{I}_{t}})]
\end{aligned}
\end{equation}
In practice, the features can be directly extracted from \emph{Reduction}. So the objective can be removed to save parameters and computation.

\subsection{Accompanying Learning}

\begin{figure}[!t]
\centering
\includegraphics[width=3.0in]{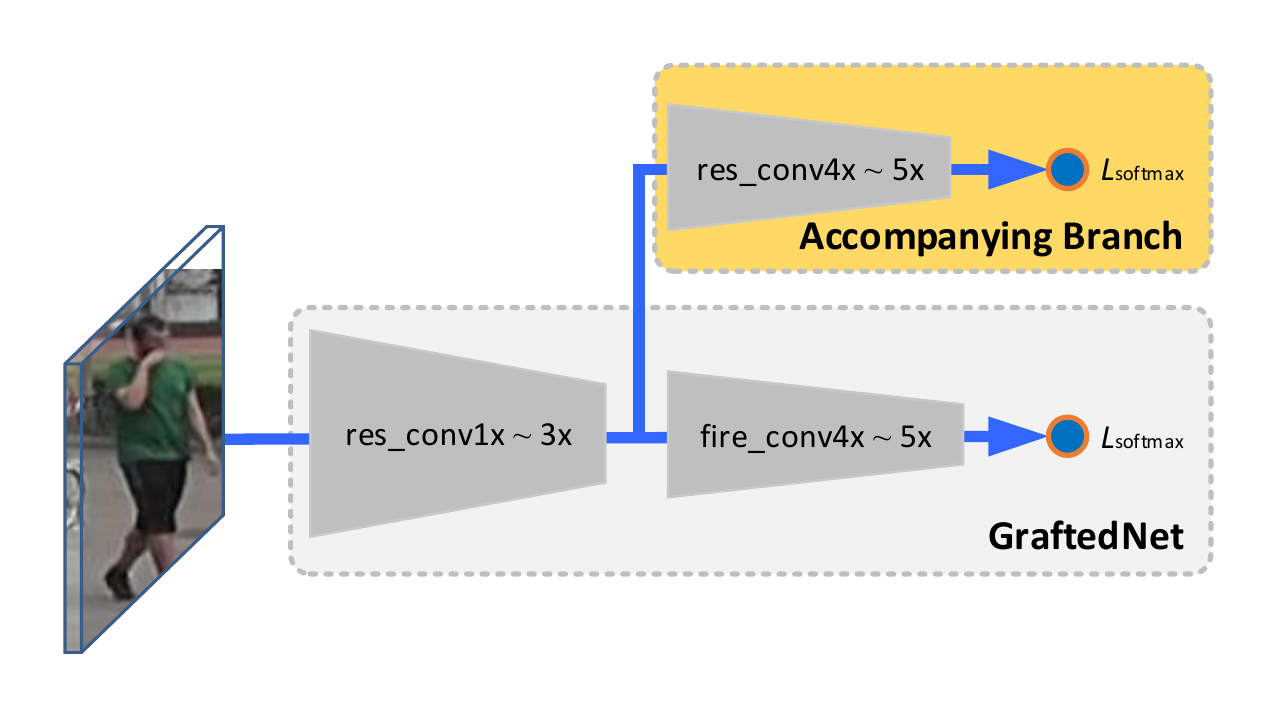}
\caption{The architecture of GraftedNet with accompanying branch.}
\label{fig4_Accompanying}
\end{figure}

Accompanying learning is simple but effective. In experiments, we find that GraftedNet shows low accuracy when we directly train it. To tackle this problem, we add another accompanying branch, which is composed of res\_conv4x and res\_conv5x of ResNet50, shown in Figure \ref{fig4_Accompanying}.

As a result, except for \emph{Rootstock}, both accompanying branch and \emph{Scion} are trained for re-ID, but accompanying branch is initialized with the ImageNet pre-trained parameter, while \emph{Scion} is initialized with random parameters. Accompanying branch and \emph{Scion}, just as a senior and a freshman, are learning the same classification task. According to the architecture of GraftedNet, the senior and the freshman have the same baseline knowledge, the output of \emph{Rootstock}. However, the senior has more prior knowledge, which comes from the ImageNets pre-trained parameter, while the freshman has only random knowledge. When they learn the task at same time, both of them can update \emph{Rootstock}. As the senior has more prior knowledge, so it can help the freshman to achieve better performance, especially when the freshman has no prior knowledge. In practice, the learning rate of \emph{Rootstock} and accompanying branch is smaller than that of \emph{Scion}. The reason is that the freshman (\emph{Scion}) has better learning ability than the seniors (\emph{Rootstock} and accompanying branch).

\section{Experiments}
\subsection{Datasets}
Experiments are conducted on three public datasets, including Market1501~\cite{DBLP:conf/iccv/ZhengSTWWT15}, DukeMTMC-reID~\cite{DBLP:conf/iccv/ZhengZY17}, CUHK03~\cite{DBLP:conf/cvpr/LiZXW14}. The Market1501 dataset is collected in Tsinghua University. It contains 32668 annotated bounding boxes of 1501 identities, among which 751 identities of 12936 images for training, and 3368 query images of 750 identities and 19732 gallery images of corresponding identities and clutter and background for testing. The DukeMTMC-reID is a subset of the DukeMTMC dataset. It consists of 16522 images of 702 identities for training and 2228 query images of the other 702 identities and 17661 gallery images (including 702 identities and 408 distractors). The CUHK03 dataset is collected from cameras in CUHK campus. There are 7365 training images of 707 identities and 1400 query images of other 700 identities and 5332 gallery images of corresponding 700 identities. For Market1501 and DukeMTMC-reID, we use the standard evaluation protocol~\cite{DBLP:conf/iccv/ZhengSTWWT15}, while we use the new training and testing protocol for CUHK03~\cite{DBLP:conf/cvpr/ZhongZCL17}.

\subsection{Implementation}
For training, we resize the input image to 384$\times$192. Data augmentation includes random cropping, horizontal flipping and random erasing ~\cite{DBLP:journals/corr/abs-1708-04896}. The parameters in \emph{Rootstock} and accompanying branch are initialized by the parameters of ImageNet pre-trained ResNet-50 model. Other parameters in GraftedNet are initialized by the `arxiver' method~\cite{DBLP:conf/cvpr/HeZRS16}. The mini-batch size of training is 32, and the examples are shuffled randomly. The SGD optimizer is used with momentum 0.9. The weight decay factor is set to 0.0005. The total training has 80 epochs. The learning rate of the parameters in \emph{Rootstock} and accompanying branch is initialized to 0.01, and the learning rate of the parameters in designed fire modules, \emph{Reduction} and \emph{Objective} are initialized to 0.1 for faster convergence. The learning rate decays by a factor of 0.1 at 40 and 60 epochs.

For testing, we average the features extracted from an original image and its horizontal flipped one as the final feature. The cosine similarity is used for evaluating. Our model is implemented on Pytorch framework. It takes about 4 hours for training on Market1501 dataset with one GTX 1080Ti GPU. To compare the performance of different methods, the two public evaluation metrics, CMC and mean Average Precision (mAP), are used. In all experiments, we use the single query mode and report the CMC at rank-1, rank-5, rank-10 and rank-20, and mAP~\cite{DBLP:conf/iccv/ZhengSTWWT15}.

\subsection{Comparison with State-of-the-art Methods}
To evaluate the performance of GraftedNet, we compare it with the state-of-the-art methods, such as IDE model~\cite{DBLP:journals/corr/ZhengYH16}, PAN~\cite{DBLP:journals/corr/ZhengZY17aa}, SVDNet~\cite{DBLP:conf/iccv/SunZDW17}, TriNet~\cite{DBLP:journals/corr/HermansBL17}, PL-Net~\cite{DBLP:journals/corr/YaoZZLT17}, DaRe~\cite{DBLP:journals/corr/abs-1805-08805}, SAG~\cite{DBLP:journals/corr/abs-1809-08556}, MLFN~\cite{DBLP:journals/corr/abs-1803-09132}, HA-CNN~\cite{DBLP:journals/corr/abs-1802-08122}, DuATM~\cite{DBLP:journals/corr/abs-1803-09937}, PCB~\cite{DBLP:conf/eccv/SunZYTW18} DeepPerson~\cite{DBLP:conf/icb/JinWLL17}, Fusion~\cite{DBLP:journals/corr/abs-1803-10630}, SphereReID~\cite{DBLP:journals/corr/abs-1807-00537}, SPreID~\cite{DBLP:conf/cvpr/KalayehBGKS18} and MGN~\cite{DBLP:conf/mm/WangYCLZ18}. Results in details are presented in Table \ref{tab1_state-of-the-arts}, where the results of light-weighted models, SqueezeNet~\cite{DBLP:journals/corr/IandolaMAHDK16}, MobileNetV2~\cite{DBLP:conf/cvpr/SandlerHZZC18}, ShuffleNet~\cite{DBLP:conf/cvpr/ZhangZLS18}, are shown separately from other methods.

\begin{table*}[!t]
\centering
\caption{Comparison with state-of-the-art methods.}
\label{tab1_state-of-the-arts}
\small
\begin{tabular}{|l|c|c|c|c|c|c|}
\hline
 & \multicolumn{2}{c|}{\textbf{Market}} & \multicolumn{2}{c|}{\textbf{Duke}} & \multicolumn{2}{c|}{\textbf{CUHK}} \\ \hline
\textbf{Methods} & \textbf{mAP} & \textbf{Rank1} & \textbf{mAP} & \textbf{Rank1} & \textbf{mAP} & \textbf{Rank1} \\ \hline
\textbf{IDE~\cite{DBLP:journals/corr/ZhengYH16}} & 50.7\% & 75.6\% & 45.0\% & 65.2\% & 19.7\% & 21.3\% \\
\textbf{PAN~\cite{DBLP:journals/corr/ZhengZY17aa}} & 63.4\% & 82.8\% & 51.5\% & 71.6\% & 34.0\% & 36.3\% \\
\textbf{SVDNet~\cite{DBLP:conf/iccv/SunZDW17}} & 62.1\% & 82.3\% & 56.8\% & 76.7\% & 37.3\% & 41.5\% \\
\textbf{TriNet~\cite{DBLP:journals/corr/HermansBL17}} & 69.1\% & 84.9\% & -- & -- & 50.7\% & 55.5\% \\
\textbf{PL-Net~\cite{DBLP:journals/corr/YaoZZLT17}}  & 69.3\% & 88.2\% & -- & -- & -- & -- \\
\textbf{DaRe(R)~\cite{DBLP:journals/corr/abs-1805-08805}} & 69.3\% & 86.4\% & 57.4\% & 75.2\% & 51.3\% & 55.1\% \\
\textbf{DaRe(De)~\cite{DBLP:journals/corr/abs-1805-08805}} & 69.9\% & 86.0\% & 56.3\% & 74.5\% & 50.1\% & 54.3\% \\
\textbf{SAG~\cite{DBLP:journals/corr/abs-1809-08556}} & 73.9\% & 90.2\% & 60.9\% & 79.9\% & -- & -- \\
\textbf{MLFN~\cite{DBLP:journals/corr/abs-1803-09132}} & 74.3\% & 90.0\% & 62.8\% & 81.0\% & 47.8\% & 52.8\% \\
\textbf{HA-CNN~\cite{DBLP:journals/corr/abs-1802-08122}} & 75.5\% & 91.2\% & 63.8\% & 80.5\% & 38.6\% & 41.7\% \\
\textbf{DuATM~\cite{DBLP:journals/corr/abs-1803-09937}} & 76.6\% & 91.4\% & 64.6\% & 81.8\% & -- & -- \\
\textbf{PCB~\cite{DBLP:conf/eccv/SunZYTW18}} & 77.4\% & 92.3\% & 66.1\% & 81.7\% & 53.2\% & 59.7\% \\
\textbf{PCB+RPP~\cite{DBLP:conf/eccv/SunZYTW18}} & 81.6\% & 93.8\% & 69.2\% & 83.3\% & 57.5\% & 63.7\% \\
\textbf{DeepPerson~\cite{DBLP:conf/icb/JinWLL17}} & 79.6\% & 92.3\% & 64.8\% & 80.9\% & -- & -- \\
\textbf{Fusion~\cite{DBLP:journals/corr/abs-1803-10630}} & 79.1\% & 92.1\% & 64.8\% & 80.4\% & -- & -- \\
\textbf{SphereReID~\cite{DBLP:journals/corr/abs-1807-00537}} & 83.6\% & 94.4\% & 68.5\% & 83.9\% & -- & -- \\
\textbf{SPreID~\cite{DBLP:conf/cvpr/KalayehBGKS18}} & 83.4\% & 93.7\% & 73.3\% & 86.0\% & -- & -- \\
\textbf{MGN~\cite{DBLP:conf/mm/WangYCLZ18}} & 86.9\% & 95.7\% & 78.4\% & 88.7\% & 66.0\% & 66.8\% \\ \hline
\textbf{SqueezeNet~\cite{DBLP:journals/corr/IandolaMAHDK16}} & 65.8\% & 85.0\% & 54.3\% & 72.4\% & 37.7\% & 42.4\% \\
\textbf{ShuffleNet~\cite{DBLP:conf/cvpr/ZhangZLS18}} & 71.6\% & 88.9\% & 60.4\% & 78.0\% & 43.5\% & 49.1\% \\
\textbf{MobileNetV2~\cite{DBLP:conf/cvpr/SandlerHZZC18}} & 73.6\% & 88.2\% & 61.0\% & 77.9\% & 52.4\% & 57.9\% \\
\textbf{GraftedNet} & 81.6\% & 93.0\% & 74.7\% & 85.3\% & 71.6\% & 76.2\% \\ \hline
\end{tabular}
\end{table*}

From Table \ref{tab1_state-of-the-arts}, we can find that our GraftedNet achieved 93.0\% in Rank-1 and 81.6\% in mAP on Market1501, achieved 85.3\% in Rank-1 and 74.7\% in mAP on DukeMTMC-reID, and achieved 76.2\% in Rank-1 and 71.6\% in mAP on CUHK03. The result achieved by our GraftedNet surpasses most of the existing methods. Comparing with the widely used baseline, the IDE model, GraftedNet exceeds it with a large margin. Comparing with part-based models, PCB+RPP, GraftedNet has a comparative performance. Although there is a gap between GraftedNet and MGN, our GraftedNet has a small model size and little computation. Comparing with light-weighted models, SqueezeNet, ShuffleNet and MobileNetV2, our model has much better performance than them.

There is a gap between our GraftedNet and MGN, mainly because MGN uses multiple independent branches and effective triplet loss to learn more discriminative features. However, our model has 4.6M parameters, while MGN has 68.8M parameters.

\begin{figure}[!t]
\centering
\includegraphics[width=3.0in]{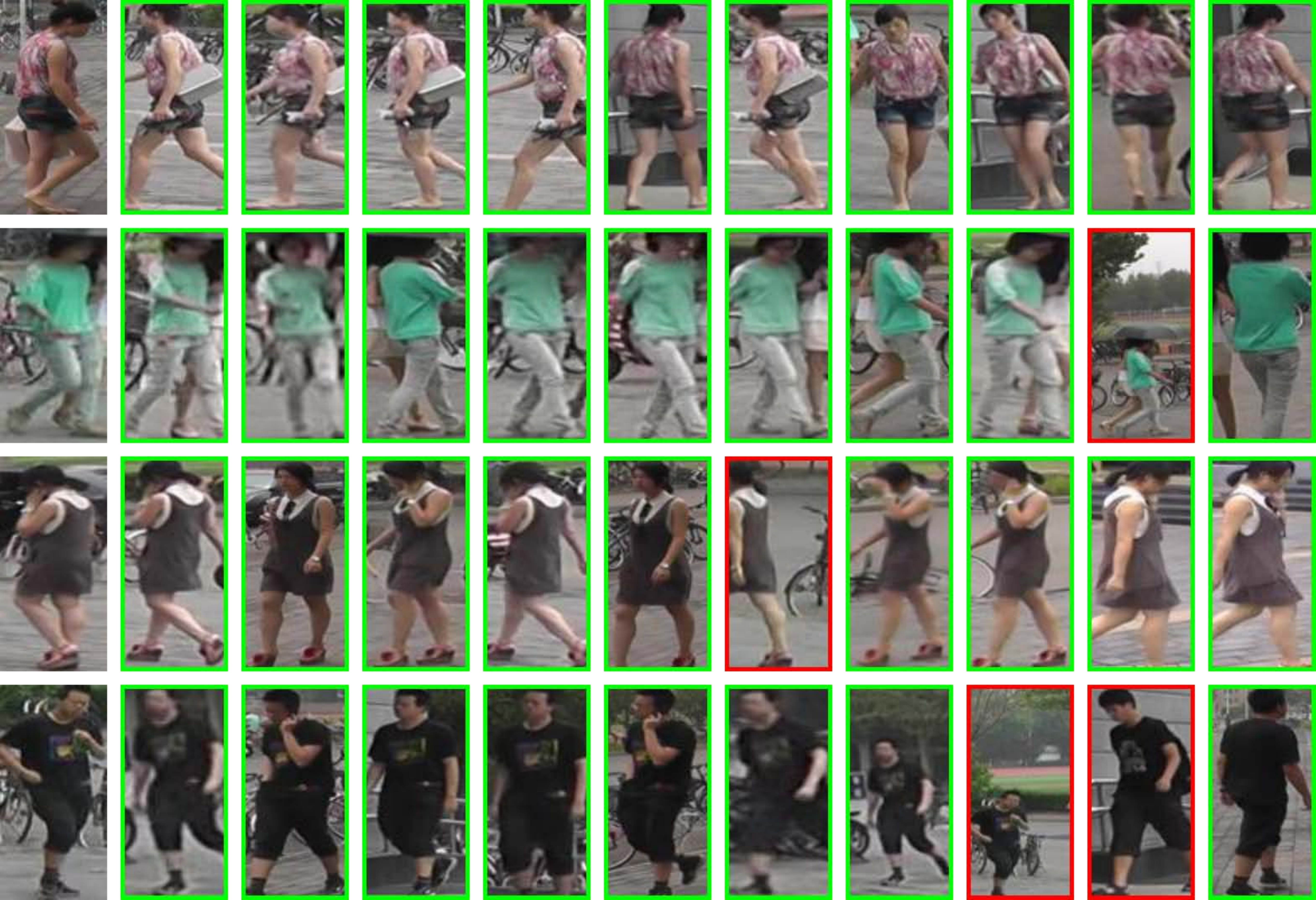}
\caption{Examples retrieved by GraftedNet on Market1501}
\label{fig5_Examples}
\end{figure}

In order to show the effect of this method intuitively, four query pedestrians are randomly selected from the Market1501 and the results of the first 10 rankings are returned, as shown in Figure \ref{fig5_Examples}. Four pedestrian images are listed in the first column, and the first 10 results are listed from the second column to the eleventh column. In the return results, the green border is the correct result and the red border is the wrong result.

From Figure \ref{fig5_Examples}, it can be found that there are large changes in perspective, posture and size between the results of the four queries and the query pedestrian images. The first query has the first 10 results with the same ID. The second query has a blurred image of pedestrian, but still returns 9 correct results. The seventh result is an error, which is an interference image in the database. The third query has similar results as the second, but the returned results are very different in perspective. The sixth result is also an interference image in the database. The last query has two incorrect results, of which the eighth result is the interference image in the database, and the ninth result is another person in database. Generally speaking, although GraftedNet just has 4.6M in model size, it has 98.34\% in Rank-10, which is a very effective accuracy for person re-ID.

\subsection{Components Analysis}
In following analysis, we define GraftedNet as our Baseline, and perform an ablation study to demonstrate the effectiveness of joint multi-level and part-based feature and the accompanying learning. Besides, we also present that the group convolution used in \emph{Reduction} can reduce the parameters greatly with a very little degradation of effectiveness.

\subsubsection{Joint Multi-level and Part-based Feature}
To further analyze the contributions of the multi-level feature and the part-based feature, we reduce them one by one. Firstly, we remove the multi-level feature, and only extract the part-based feature from fire\_conv5c. We note this as `-MF'. Then we remove the part-based feature, and only extract features from fire\_conv4f, fire\_conv5a, fire\_conv5b and fire\_conv5c and concatenate as the final feature. We note this as `-PF'. Finally, we remove both the multi-level feature and the part-based feature, and only extract a global feature from fire\_conv5c. We note this as `-MF-PF'. The evaluated results are showed in Table \ref{tab2_multilevel_part}. Note that group convolution is used in \emph{Reduction} and accompanying learning is also used for fair comparison.
\begin{table*}[!t]
\centering
\caption{Component analysis on Market1501.}
\label{tab2_multilevel_part}
\small
\begin{tabular}{|l|r|r|r|r|r|r|}
\hline
\textbf{Methods} & \textbf{mAP} & \textbf{Rank1} & \textbf{Rank5} & \textbf{Rank10} & \textbf{Rank20} & \textbf{Params} \\ \hline
\textbf{GraftedNet} & 81.58\% & 93.02\% & 97.27\% & 98.34\% & 98.81\% & -- \\
\textbf{GraftedNet(-MF)} & 74.27\% & 89.88\% & 96.35\% & 97.71\% & 98.49\% & -- \\
\textbf{GraftedNet(-PF)} & 72.14\% & 88.00\% & 94.83\% & 96.76\% & 97.83\% & -- \\
\textbf{GraftedNet(-MF-PF)} & 67.19\% & 85.07\% & 94.45\% & 96.44\% & 97.83\% & -- \\
%\textbf{GraftedNet} & 81.58\% & 93.02\% & 97.27\% & 98.34\% & 98.81\% & -- \\
\textbf{GraftedNet(-AL)} & 78.28\% & 91.24\% & 96.62\% & 97.86\% & 98.52\% & -- \\ \hline
\textbf{GraftedNet (g=8)} & 81.58\% & 93.02\% & 97.27\% & 98.34\% & 98.81\% & 0.218M \\
\textbf{GraftedNet (g=4)} & 81.63\% & 93.14\% & 97.27\% & 98.34\% & 98.90\% & 0.431M \\
\textbf{GraftedNet (g=1)} & 82.01\% & 92.87\% & 97.42\% & 98.40\% & 98.84\% & 1.709M \\ \hline
\end{tabular}
\end{table*}

From Table \ref{tab2_multilevel_part}, we can find that mAP and Rank-1 have dropped from 81.58\% and 93.02\% to 74.27\% and 89.88\% without the multi-level feature. When the part-based feature is removed, mAP and Rank-1 have a drop of 9.44\% and 5.02\% respectively. When both removed, there is a huge drop of 14.39\% in mAP and 7.95\% in Rank-1. These results demonstrate the contribution of joint multi-level feature and part-based feature.

\subsubsection{Accompanying Learning}
In order to further analysis the influence of the accompanying learning, we compare the results of GraftedNet with/without accompanying learning. Table \ref{tab2_multilevel_part} shows the results, where `-AL' means training without accompanying learning. Note that the parameters in \emph{Rootstock} are initialized by the ImageNet pre-trained parameters, while other parameters are initialized by `arxiver' method.

%\begin{table*}[!t]
%\centering
%\caption{Results of Accompanying Learning on Market1501.}
%\label{tab3_Accompanying}
%\small
%\begin{tabular}{|l|c|c|c|c|c|}
%\hline
%\textbf{Methods} & \textbf{mAP} & \textbf{Rank1} & \textbf{Rank5} & \textbf{Rank10} & \textbf{Rank20} \\ \hline
%\textbf{GraftedNet} & 81.58\% & 93.02\% & 97.27\% & 98.34\% & 98.81\% \\
%\textbf{GraftedNet(-AL)} & 78.28\% & 91.24\% & 96.62\% & 97.86\% & 98.52\% \\ \hline
%\end{tabular}
%\end{table*}
From Table \ref{tab2_multilevel_part}, we can find that GraftedNet without accompanying learning has a drop of 3.30\% in mAP and 1.78\% in Rank-1. This comparison demonstrates the effectiveness of accompanying learning for GraftedNet.

\subsubsection{Group Convolution}
\label{group}
In experiments, we find that group convolution can further reduce the parameters. For \emph{Reduction}, it has been repeated 9 times when joint multi-level and part-based feature is extracted. So it has 1.709M parameters in total. To reduce the parameters, we explore different group size to get a balance between accuracy and the number of parameters. The results with the number of group $g$ ($g=\{8,4,1\}$) are shown in Table \ref{tab2_multilevel_part}, where `Params' means the number of parameters only in \emph{Reduction}.

%\begin{table*}[!t]
%\centering
%\caption{Results of Group Convolution on Market1501.}
%\label{tab4_group_conv}
%\small
%\begin{tabular}{|l|c|c|c|c|c|r|}
%\hline
%\textbf{Methods} & \textbf{mAP} & \textbf{Rank1} & \textbf{Rank5} & \textbf{Rank10} & \textbf{Rank20} & \textbf{Params} \\ \hline
%\textbf{GraftedNet (g=8)} & 81.58\% & 93.02\% & 97.27\% & 98.34\% & 98.81\% & 0.218M \\
%\textbf{GraftedNet (g=4)} & 81.63\% & 93.14\% & 97.27\% & 98.34\% & 98.90\% & 0.431M \\
%\textbf{GraftedNet (g=1)} & 82.01\% & 92.87\% & 97.42\% & 98.40\% & 98.84\% & 1.709M \\ \hline
%\end{tabular}
%\end{table*}

From Table \ref{tab2_multilevel_part}, we can find that when GraftedNet of $g=8$ reduces the size from 1.709M to 0.218M, with several times reduction. Meanwhile, there is a very little drop of the performance. As a result, group convolution is an effective strategy for compressing model while maintaining accuracy.

\subsection{Memory and Computation Analysis}
In order to compare with the existing light-weighted models, such as SqueezeNet~\cite{DBLP:journals/corr/IandolaMAHDK16}\footnote{https://github.com/pytorch/vision/tree/master/torchvision/models}, MobileNetV2~\cite{DBLP:conf/cvpr/SandlerHZZC18}\footnote{https://github.com/tonylins/pytorch-mobilenet-v2}, ShuffleNet~\cite{DBLP:conf/cvpr/ZhangZLS18}\footnote{https://github.com/KaiyangZhou/deep-person-reid/blob/master/torchreid/models/}, we presents the results of these models with fine-tuning on Market1501 in Table \ref{tab5_light_weighted}, where `Params' means the number of parameters in the corresponding models, and `Times' means the computation cost (mini-seconds) of each image in testing. Besides, we also present the performance of original ResNet50. The comparison is conducted on a workstation with 16 Intel Xeon CPU (3.50GHz).
\begin{table}[!t]
\centering
\caption{Memory and computation analysis on Market1501.}
\label{tab5_light_weighted}
\small
\begin{tabular}{|l|r|r|r|r|}
\hline
\textbf{Methods}     & \textbf{Rank-1}  & \textbf{mAP}     & \textbf{Params} & \textbf{Time}     \\ \hline
\textbf{SqueezeNet}  & 85.04\% & 65.82\% & 1.05M    & 44.92ms  \\
\textbf{ShuffleNet}  & 88.87\% & 71.63\% & 1.34M    & 49.33ms  \\
\textbf{MobileNetV2} & 88.15\% & 73.64\% & 2.75M    & 106.17ms \\
\textbf{ResNet50}    & 91.83\% & 78.67\% & 24.23M   & 408.64ms \\
\textbf{GraftedNet}  & 93.02\% & 81.58\% & 4.60M    & 268.57ms \\ \hline
\end{tabular}
\end{table}

From Table \ref{tab5_light_weighted}, we can find that GraftedNet sacrifices amount of computation and storage costs to achieve better performance, comparing with the three light-weighted models. However, compared with original ResNet-50, GraftedNet has better performance with less computation and smaller model size.

\section{Conclusion}
In this paper, we propose a novel light-weighted and high-accuracy grafted network (GraftedNet), which achieves better performance than the original ResNet-50 model, with only 4.6M parameters. A joint multi-level and part-based feature is proposed for describing each person image. To train the network, an accompanying learning branch is proposed and can be removed in testing for saving parameters. The effectiveness of GraftedNet has been evaluated and related components are compared and analyzed on the public datasets. Compared with existing light-weighted networks, our GraftedNet achieves much better performance. In the further, we plan to add attention mechanism to our GratfedNet, such as the Squeeze-and-Excitation (SE) module~\cite{DBLP:conf/cvpr/HuSS18}, to further improve the performance.

\section*{References}
\bibliographystyle{elsarticle-num}
\bibliography{GraftedNet}

\begin{thebibliography}{10}
\expandafter\ifx\csname url\endcsname\relax
  \def\url#1{\texttt{#1}}\fi
\expandafter\ifx\csname urlprefix\endcsname\relax\def\urlprefix{URL }\fi
\expandafter\ifx\csname href\endcsname\relax
  \def\href#1#2{#2} \def\path#1{#1}\fi

\bibitem{DBLP:conf/cvpr/HeZRS16}
K.~He, X.~Zhang, S.~Ren, J.~Sun,
  \href{https://doi.org/10.1109/CVPR.2016.90}{Deep residual learning for image
  recognition}, in: 2016 {IEEE} Conference on Computer Vision and Pattern
  Recognition, {CVPR} 2016, Las Vegas, NV, USA, June 27-30, 2016, 2016, pp.
  770--778.
\newblock \href {https://doi.org/10.1109/CVPR.2016.90}
  {\path{doi:10.1109/CVPR.2016.90}}.
\newline\urlprefix\url{https://doi.org/10.1109/CVPR.2016.90}

\bibitem{DBLP:journals/corr/IandolaMAHDK16}
F.~N. Iandola, M.~W. Moskewicz, K.~Ashraf, S.~Han, W.~J. Dally, K.~Keutzer,
  \href{http://arxiv.org/abs/1602.07360}{Squeezenet: Alexnet-level accuracy
  with 50x fewer parameters and {\textless}1mb model size}, CoRR abs/1602.07360
  (2016).
\newblock \href {http://arxiv.org/abs/1602.07360} {\path{arXiv:1602.07360}}.
\newline\urlprefix\url{http://arxiv.org/abs/1602.07360}

\bibitem{DBLP:journals/corr/HowardZCKWWAA17}
A.~G. Howard, M.~Zhu, B.~Chen, D.~Kalenichenko, W.~Wang, T.~Weyand,
  M.~Andreetto, H.~Adam, \href{http://arxiv.org/abs/1704.04861}{Mobilenets:
  Efficient convolutional neural networks for mobile vision applications}, CoRR
  abs/1704.04861 (2017).
\newblock \href {http://arxiv.org/abs/1704.04861} {\path{arXiv:1704.04861}}.
\newline\urlprefix\url{http://arxiv.org/abs/1704.04861}

\bibitem{DBLP:conf/cvpr/ZhangZLS18}
X.~Zhang, X.~Zhou, M.~Lin, J.~Sun,
  \href{http://openaccess.thecvf.com/content\_cvpr\_2018/html/Zhang\_ShuffleNet\_An\_Extremely\_CVPR\_2018\_paper.html}{Shufflenet:
  An extremely efficient convolutional neural network for mobile devices}, in:
  2018 {IEEE} Conference on Computer Vision and Pattern Recognition, {CVPR}
  2018, Salt Lake City, UT, USA, June 18-22, 2018, 2018, pp. 6848--6856.
\newblock \href {https://doi.org/10.1109/CVPR.2018.00716}
  {\path{doi:10.1109/CVPR.2018.00716}}.
\newline\urlprefix\url{http://openaccess.thecvf.com/content\_cvpr\_2018/html/Zhang\_ShuffleNet\_An\_Extremely\_CVPR\_2018\_paper.html}

\bibitem{DBLP:journals/corr/SimonyanZ14a}
K.~Simonyan, A.~Zisserman, \href{http://arxiv.org/abs/1409.1556}{Very deep
  convolutional networks for large-scale image recognition}, CoRR abs/1409.1556
  (2014).
\newblock \href {http://arxiv.org/abs/1409.1556} {\path{arXiv:1409.1556}}.
\newline\urlprefix\url{http://arxiv.org/abs/1409.1556}

\bibitem{DBLP:conf/nips/KrizhevskySH12}
A.~Krizhevsky, I.~Sutskever, G.~E. Hinton,
  \href{http://papers.nips.cc/paper/4824-imagenet-classification-with-deep-convolutional-neural-networks}{Imagenet
  classification with deep convolutional neural networks}, in: Advances in
  Neural Information Processing Systems 25: 26th Annual Conference on Neural
  Information Processing Systems 2012. Proceedings of a meeting held December
  3-6, 2012, Lake Tahoe, Nevada, United States., 2012, pp. 1106--1114.
\newline\urlprefix\url{http://papers.nips.cc/paper/4824-imagenet-classification-with-deep-convolutional-neural-networks}

\bibitem{DBLP:conf/cvpr/SandlerHZZC18}
M.~Sandler, A.~G. Howard, M.~Zhu, A.~Zhmoginov, L.~Chen,
  \href{http://openaccess.thecvf.com/content\_cvpr\_2018/html/Sandler\_MobileNetV2\_Inverted\_Residuals\_CVPR\_2018\_paper.html}{Mobilenetv2:
  Inverted residuals and linear bottlenecks}, in: 2018 {IEEE} Conference on
  Computer Vision and Pattern Recognition, {CVPR} 2018, Salt Lake City, UT,
  USA, June 18-22, 2018, 2018, pp. 4510--4520.
\newblock \href {https://doi.org/10.1109/CVPR.2018.00474}
  {\path{doi:10.1109/CVPR.2018.00474}}.
\newline\urlprefix\url{http://openaccess.thecvf.com/content\_cvpr\_2018/html/Sandler\_MobileNetV2\_Inverted\_Residuals\_CVPR\_2018\_paper.html}

\bibitem{DBLP:conf/eccv/MaZZS18}
N.~Ma, X.~Zhang, H.~Zheng, J.~Sun,
  \href{https://doi.org/10.1007/978-3-030-01264-9\_8}{Shufflenet {V2:}
  practical guidelines for efficient {CNN} architecture design}, in: Computer
  Vision - {ECCV} 2018 - 15th European Conference, Munich, Germany, September
  8-14, 2018, Proceedings, Part {XIV}, 2018, pp. 122--138.
\newblock \href {https://doi.org/10.1007/978-3-030-01264-9\_8}
  {\path{doi:10.1007/978-3-030-01264-9\_8}}.
\newline\urlprefix\url{https://doi.org/10.1007/978-3-030-01264-9\_8}

\bibitem{DBLP:journals/corr/HintonVD15}
G.~E. Hinton, O.~Vinyals, J.~Dean,
  \href{http://arxiv.org/abs/1503.02531}{Distilling the knowledge in a neural
  network}, CoRR abs/1503.02531 (2015).
\newblock \href {http://arxiv.org/abs/1503.02531} {\path{arXiv:1503.02531}}.
\newline\urlprefix\url{http://arxiv.org/abs/1503.02531}

\bibitem{DBLP:journals/corr/abs-1902-00643}
S.~Zhang, J.~Li, B.~Zhang, \href{http://arxiv.org/abs/1902.00643}{Pairwise
  teacher-student network for semi-supervised hashing}, CoRR abs/1902.00643
  (2019).
\newblock \href {http://arxiv.org/abs/1902.00643} {\path{arXiv:1902.00643}}.
\newline\urlprefix\url{http://arxiv.org/abs/1902.00643}

\bibitem{DBLP:conf/cvpr/ZhangXHL18}
Y.~Zhang, T.~Xiang, T.~M. Hospedales, H.~Lu,
  \href{http://openaccess.thecvf.com/content\_cvpr\_2018/html/Zhang\_Deep\_Mutual\_Learning\_CVPR\_2018\_paper.html}{Deep
  mutual learning}, in: 2018 {IEEE} Conference on Computer Vision and Pattern
  Recognition, {CVPR} 2018, Salt Lake City, UT, USA, June 18-22, 2018, 2018,
  pp. 4320--4328.
\newblock \href {https://doi.org/10.1109/CVPR.2018.00454}
  {\path{doi:10.1109/CVPR.2018.00454}}.
\newline\urlprefix\url{http://openaccess.thecvf.com/content\_cvpr\_2018/html/Zhang\_Deep\_Mutual\_Learning\_CVPR\_2018\_paper.html}

\bibitem{DBLP:conf/mm/WangYCLZ18}
G.~Wang, Y.~Yuan, X.~Chen, J.~Li, X.~Zhou,
  \href{https://doi.org/10.1145/3240508.3240552}{Learning discriminative
  features with multiple granularities for person re-identification}, in: 2018
  {ACM} Multimedia Conference on Multimedia Conference, {MM} 2018, Seoul,
  Republic of Korea, October 22-26, 2018, 2018, pp. 274--282.
\newblock \href {https://doi.org/10.1145/3240508.3240552}
  {\path{doi:10.1145/3240508.3240552}}.
\newline\urlprefix\url{https://doi.org/10.1145/3240508.3240552}

\bibitem{DBLP:conf/eccv/SunZYTW18}
Y.~Sun, L.~Zheng, Y.~Yang, Q.~Tian, S.~Wang,
  \href{https://doi.org/10.1007/978-3-030-01225-0\_30}{Beyond part models:
  Person retrieval with refined part pooling (and {A} strong convolutional
  baseline)}, in: Computer Vision - {ECCV} 2018 - 15th European Conference,
  Munich, Germany, September 8-14, 2018, Proceedings, Part {IV}, 2018, pp.
  501--518.
\newblock \href {https://doi.org/10.1007/978-3-030-01225-0\_30}
  {\path{doi:10.1007/978-3-030-01225-0\_30}}.
\newline\urlprefix\url{https://doi.org/10.1007/978-3-030-01225-0\_30}

\bibitem{DBLP:conf/cvpr/SzegedyLJSRAEVR15}
C.~Szegedy, W.~Liu, Y.~Jia, P.~Sermanet, S.~E. Reed, D.~Anguelov, D.~Erhan,
  V.~Vanhoucke, A.~Rabinovich,
  \href{https://doi.org/10.1109/CVPR.2015.7298594}{Going deeper with
  convolutions}, in: {IEEE} Conference on Computer Vision and Pattern
  Recognition, {CVPR} 2015, Boston, MA, USA, June 7-12, 2015, 2015, pp. 1--9.
\newblock \href {https://doi.org/10.1109/CVPR.2015.7298594}
  {\path{doi:10.1109/CVPR.2015.7298594}}.
\newline\urlprefix\url{https://doi.org/10.1109/CVPR.2015.7298594}

\bibitem{DBLP:conf/iccv/ZhengSTWWT15}
L.~Zheng, L.~Shen, L.~Tian, S.~Wang, J.~Wang, Q.~Tian,
  \href{https://doi.org/10.1109/ICCV.2015.133}{Scalable person
  re-identification: {A} benchmark}, in: 2015 {IEEE} International Conference
  on Computer Vision, {ICCV} 2015, Santiago, Chile, December 7-13, 2015, 2015,
  pp. 1116--1124.
\newblock \href {https://doi.org/10.1109/ICCV.2015.133}
  {\path{doi:10.1109/ICCV.2015.133}}.
\newline\urlprefix\url{https://doi.org/10.1109/ICCV.2015.133}

\bibitem{DBLP:conf/iccv/ZhengZY17}
Z.~Zheng, L.~Zheng, Y.~Yang,
  \href{https://doi.org/10.1109/ICCV.2017.405}{Unlabeled samples generated by
  {GAN} improve the person re-identification baseline in vitro}, in: {IEEE}
  International Conference on Computer Vision, {ICCV} 2017, Venice, Italy,
  October 22-29, 2017, 2017, pp. 3774--3782.
\newblock \href {https://doi.org/10.1109/ICCV.2017.405}
  {\path{doi:10.1109/ICCV.2017.405}}.
\newline\urlprefix\url{https://doi.org/10.1109/ICCV.2017.405}

\bibitem{DBLP:conf/cvpr/LiZXW14}
W.~Li, R.~Zhao, T.~Xiao, X.~Wang,
  \href{https://doi.org/10.1109/CVPR.2014.27}{Deepreid: Deep filter pairing
  neural network for person re-identification}, in: 2014 {IEEE} Conference on
  Computer Vision and Pattern Recognition, {CVPR} 2014, Columbus, OH, USA, June
  23-28, 2014, 2014, pp. 152--159.
\newblock \href {https://doi.org/10.1109/CVPR.2014.27}
  {\path{doi:10.1109/CVPR.2014.27}}.
\newline\urlprefix\url{https://doi.org/10.1109/CVPR.2014.27}

\bibitem{DBLP:conf/cvpr/ZhongZCL17}
Z.~Zhong, L.~Zheng, D.~Cao, S.~Li,
  \href{https://doi.org/10.1109/CVPR.2017.389}{Re-ranking person
  re-identification with k-reciprocal encoding}, in: 2017 {IEEE} Conference on
  Computer Vision and Pattern Recognition, {CVPR} 2017, Honolulu, HI, USA, July
  21-26, 2017, 2017, pp. 3652--3661.
\newblock \href {https://doi.org/10.1109/CVPR.2017.389}
  {\path{doi:10.1109/CVPR.2017.389}}.
\newline\urlprefix\url{https://doi.org/10.1109/CVPR.2017.389}

\bibitem{DBLP:journals/corr/abs-1708-04896}
Z.~Zhong, L.~Zheng, G.~Kang, S.~Li, Y.~Yang,
  \href{http://arxiv.org/abs/1708.04896}{Random erasing data augmentation},
  CoRR abs/1708.04896 (2017).
\newblock \href {http://arxiv.org/abs/1708.04896} {\path{arXiv:1708.04896}}.
\newline\urlprefix\url{http://arxiv.org/abs/1708.04896}

\bibitem{DBLP:journals/corr/ZhengYH16}
L.~Zheng, Y.~Yang, A.~G. Hauptmann,
  \href{http://arxiv.org/abs/1610.02984}{Person re-identification: Past,
  present and future}, CoRR abs/1610.02984 (2016).
\newblock \href {http://arxiv.org/abs/1610.02984} {\path{arXiv:1610.02984}}.
\newline\urlprefix\url{http://arxiv.org/abs/1610.02984}

\bibitem{DBLP:journals/corr/ZhengZY17aa}
Z.~Zheng, L.~Zheng, Y.~Yang, \href{http://arxiv.org/abs/1707.00408}{Pedestrian
  alignment network for large-scale person re-identification}, CoRR
  abs/1707.00408 (2017).
\newblock \href {http://arxiv.org/abs/1707.00408} {\path{arXiv:1707.00408}}.
\newline\urlprefix\url{http://arxiv.org/abs/1707.00408}

\bibitem{DBLP:conf/iccv/SunZDW17}
Y.~Sun, L.~Zheng, W.~Deng, S.~Wang,
  \href{https://doi.org/10.1109/ICCV.2017.410}{Svdnet for pedestrian
  retrieval}, in: {IEEE} International Conference on Computer Vision, {ICCV}
  2017, Venice, Italy, October 22-29, 2017, 2017, pp. 3820--3828.
\newblock \href {https://doi.org/10.1109/ICCV.2017.410}
  {\path{doi:10.1109/ICCV.2017.410}}.
\newline\urlprefix\url{https://doi.org/10.1109/ICCV.2017.410}

\bibitem{DBLP:journals/corr/HermansBL17}
A.~Hermans, L.~Beyer, B.~Leibe, \href{http://arxiv.org/abs/1703.07737}{In
  defense of the triplet loss for person re-identification}, CoRR
  abs/1703.07737 (2017).
\newblock \href {http://arxiv.org/abs/1703.07737} {\path{arXiv:1703.07737}}.
\newline\urlprefix\url{http://arxiv.org/abs/1703.07737}

\bibitem{DBLP:journals/corr/YaoZZLT17}
H.~Yao, S.~Zhang, Y.~Zhang, J.~Li, Q.~Tian,
  \href{http://arxiv.org/abs/1707.00798}{Deep representation learning with part
  loss for person re-identification}, CoRR abs/1707.00798 (2017).
\newblock \href {http://arxiv.org/abs/1707.00798} {\path{arXiv:1707.00798}}.
\newline\urlprefix\url{http://arxiv.org/abs/1707.00798}

\bibitem{DBLP:journals/corr/abs-1805-08805}
Y.~Wang, L.~Wang, Y.~You, X.~Zou, V.~Chen, S.~Li, G.~Huang, B.~Hariharan, K.~Q.
  Weinberger, \href{http://arxiv.org/abs/1805.08805}{Resource aware person
  re-identification across multiple resolutions}, CoRR abs/1805.08805 (2018).
\newblock \href {http://arxiv.org/abs/1805.08805} {\path{arXiv:1805.08805}}.
\newline\urlprefix\url{http://arxiv.org/abs/1805.08805}

\bibitem{DBLP:journals/corr/abs-1809-08556}
J.~Ainam, K.~Qin, G.~Liu, \href{http://arxiv.org/abs/1809.08556}{Self attention
  grid for person re-identification}, CoRR abs/1809.08556 (2018).
\newblock \href {http://arxiv.org/abs/1809.08556} {\path{arXiv:1809.08556}}.
\newline\urlprefix\url{http://arxiv.org/abs/1809.08556}

\bibitem{DBLP:journals/corr/abs-1803-09132}
X.~Chang, T.~M. Hospedales, T.~Xiang,
  \href{http://arxiv.org/abs/1803.09132}{Multi-level factorisation net for
  person re-identification}, CoRR abs/1803.09132 (2018).
\newblock \href {http://arxiv.org/abs/1803.09132} {\path{arXiv:1803.09132}}.
\newline\urlprefix\url{http://arxiv.org/abs/1803.09132}

\bibitem{DBLP:journals/corr/abs-1802-08122}
W.~Li, X.~Zhu, S.~Gong, \href{http://arxiv.org/abs/1802.08122}{Harmonious
  attention network for person re-identification}, CoRR abs/1802.08122 (2018).
\newblock \href {http://arxiv.org/abs/1802.08122} {\path{arXiv:1802.08122}}.
\newline\urlprefix\url{http://arxiv.org/abs/1802.08122}

\bibitem{DBLP:journals/corr/abs-1803-09937}
J.~Si, H.~Zhang, C.~Li, J.~Kuen, X.~Kong, A.~C. Kot, G.~Wang,
  \href{http://arxiv.org/abs/1803.09937}{Dual attention matching network for
  context-aware feature sequence based person re-identification}, CoRR
  abs/1803.09937 (2018).
\newblock \href {http://arxiv.org/abs/1803.09937} {\path{arXiv:1803.09937}}.
\newline\urlprefix\url{http://arxiv.org/abs/1803.09937}

\bibitem{DBLP:conf/icb/JinWLL17}
H.~Jin, X.~Wang, S.~Liao, S.~Z. Li,
  \href{https://doi.org/10.1109/BTAS.2017.8272706}{Deep person
  re-identification with improved embedding and efficient training}, in: 2017
  {IEEE} International Joint Conference on Biometrics, {IJCB} 2017, Denver, CO,
  USA, October 1-4, 2017, 2017, pp. 261--267.
\newblock \href {https://doi.org/10.1109/BTAS.2017.8272706}
  {\path{doi:10.1109/BTAS.2017.8272706}}.
\newline\urlprefix\url{https://doi.org/10.1109/BTAS.2017.8272706}

\bibitem{DBLP:journals/corr/abs-1803-10630}
J.~Johnson, S.~Yasugi, Y.~Sugino, S.~Pranata, S.~Shen,
  \href{http://arxiv.org/abs/1803.10630}{Person re-identification with fusion
  of hand-crafted and deep pose-based body region features}, CoRR
  abs/1803.10630 (2018).
\newblock \href {http://arxiv.org/abs/1803.10630} {\path{arXiv:1803.10630}}.
\newline\urlprefix\url{http://arxiv.org/abs/1803.10630}

\bibitem{DBLP:journals/corr/abs-1807-00537}
X.~Fan, W.~Jiang, H.~Luo, M.~Fei,
  \href{http://arxiv.org/abs/1807.00537}{Spherereid: Deep hypersphere manifold
  embedding for person re-identification}, CoRR abs/1807.00537 (2018).
\newblock \href {http://arxiv.org/abs/1807.00537} {\path{arXiv:1807.00537}}.
\newline\urlprefix\url{http://arxiv.org/abs/1807.00537}

\bibitem{DBLP:conf/cvpr/KalayehBGKS18}
M.~M. Kalayeh, E.~Basaran, M.~G{\"{o}}kmen, M.~E. Kamasak, M.~Shah,
  \href{http://openaccess.thecvf.com/content\_cvpr\_2018/html/Kalayeh\_Human\_Semantic\_Parsing\_CVPR\_2018\_paper.html}{Human
  semantic parsing for person re-identification}, in: 2018 {IEEE} Conference on
  Computer Vision and Pattern Recognition, {CVPR} 2018, Salt Lake City, UT,
  USA, June 18-22, 2018, 2018, pp. 1062--1071.
\newblock \href {https://doi.org/10.1109/CVPR.2018.00117}
  {\path{doi:10.1109/CVPR.2018.00117}}.
\newline\urlprefix\url{http://openaccess.thecvf.com/content\_cvpr\_2018/html/Kalayeh\_Human\_Semantic\_Parsing\_CVPR\_2018\_paper.html}

\bibitem{DBLP:conf/cvpr/HuSS18}
J.~Hu, L.~Shen, G.~Sun,
  \href{http://openaccess.thecvf.com/content\_cvpr\_2018/html/Hu\_Squeeze-and-Excitation\_Networks\_CVPR\_2018\_paper.html}{Squeeze-and-excitation
  networks}, in: 2018 {IEEE} Conference on Computer Vision and Pattern
  Recognition, {CVPR} 2018, Salt Lake City, UT, USA, June 18-22, 2018, 2018,
  pp. 7132--7141.
\newblock \href {https://doi.org/10.1109/CVPR.2018.00745}
  {\path{doi:10.1109/CVPR.2018.00745}}.
\newline\urlprefix\url{http://openaccess.thecvf.com/content\_cvpr\_2018/html/Hu\_Squeeze-and-Excitation\_Networks\_CVPR\_2018\_paper.html}

\end{thebibliography}

\end{document}